**Evaluating the performance of personal, social, health-related, biomarker and genetic data for predicting an individual's future health using machine learning: A longitudinal analysis.**

Mark A Green

Geographic Data Science Lab, Department of Geography & Planning, Roxby Building, University of Liverpool, Liverpool, L69 7ZT, UK

Email: Mark.green@liverpool.ac.uk . Tel: +44 151 794 284.

**Abstract**

As we gain access to a greater depth and range of health-related information about individuals, three questions arise: (1) Can we build better models to predict individual-level risk of ill health? (2) How much data do we need to effectively predict ill health? (3) Are new methods required to process the added complexity that new forms of data bring? The aim of the study is to apply a machine learning approach to identify the relative contribution of personal, social, health-related, biomarker and genetic data as predictors of future health in individuals. Using longitudinal data from 6830 individuals in the UK from Understanding Society (2010-12 to 2015-17), the study compares the predictive performance of five types of measures: personal (e.g. age, sex), social (e.g. occupation, education), health-related (e.g. body weight, grip strength), biomarker (e.g. cholesterol, hormones) and genetic single nucleotide polymorphisms (SNPs). The predicted outcome variable was limiting long-term illness one and five years from baseline. Two machine learning approaches were used to build predictive models: deep learning via neural networks and XGBoost (gradient boosting decision trees). Model fit was compared to traditional logistic regression models. Results found that health-related measures had the strongest prediction of future health status, with genetic data performing poorly. Machine learning models only offered marginal improvements in model accuracy when compared to logistic regression models, but also performed well on other metrics e.g. neural networks were best on AUC and XGBoost on precision. Age was consistently the strongest predictor, with some social (e.g. job permanent, job security), health-related (e.g. walk pace, physicality of work) and biomarkers (e.g. dehydroepiandrosterone sulphate, insulin-like-growth factor 1) also important. The study suggests that increasing complexity of data and methods does not necessarily translate to improved understanding of the determinants of health or performance of predictive models of ill health.

# 1 Introduction

Health research is often concerned with the identification of risk factors which influence the likelihood of disease. While this endeavour has produced great insight into how social and biological factors influence health, there has been limited success towards being able to discriminate between cases and non-cases of disease(s) among individuals (Green et al., 2017; Merlo, 2014). Research has long distinguished between how the causes of disease differ between the individual- and population-levels (Rose, 1985). For example, smoking may account for upwards of 90% of the variation at the population level, but less than 10% between individuals (Davey Smith, 2011). Poor prediction of an individual's risk of ill health may contribute to ineffective targeting and treatment options.

We are increasingly collecting greater quantities and diversity of data than ever before (Riley et al., 2016; Timmins et al., 2018), allowing us to build intricate descriptors of an individual's social and biological circumstances. Social surveys contain hundreds to thousands of variables on personal, social and health-related characteristics of participants, and are recently expanding to include biomarker data through blood samples and genetics. The explosion in data availability has yet to translate to clinical risk prediction models. For example, a review of cardiovascular risk prediction found that the median number of predictors was only seven (Damen et al., 2016), and other reviews demonstrate that few studies use more than 100 (Fahey et al., 2018; B. A. Goldstein et al., 2017; Steyerberg et al., 2018). As we gain access to a greater depth of information about individuals, two questions arise: (1) Can we build better models to predict individual-level risk of ill health? (2) How much data do we need to effectively predict ill health? Added complexity might help support personalised medicine narratives, however we require newer approaches for handling this complexity that are often missing from individual risk prediction models (B. A. Goldstein et al., 2017).

The potential of machine learning to revolutionise scientific study across multiple disciplines has been touted by many commentators (Beam & Kohane, 2018; Jordan & Mitchell, 2015). Despite successes within the field of computer science, there have been few applications across the social sciences. Within health-related fields, the main applications of machine learning have been for (i) understanding the genetic drivers of disease in Genome-Wide Association Studies (GWAS) (e.g. (Yin et al., 2019)), and (ii) processing image data to predict health outcomes (e.g. (Gulshan et al., 2016)). There has been less consideration for how machine learning approaches feed into the analysis of health and social surveys. Machine learning offers new opportunities to process data with high dimensionality making them potentially adept at analysing the rich detail collected across thousands of variables in surveys. The iterative 'learning' of data insights by algorithms may also detect underlying patterns not seen by humans producing new insights into relationships (Esteva et al., 2019). As such, they may help to bridge the gap between complexity of data and prediction modelling.

The aim of the study is to apply a machine learning approach to identify the relative contribution of personal, social, health-related, biomarker and genetic data as predictors of future health in individuals.

**2 Material and Methods**

**2.1 Data**

Wave 2 of 'Understanding Society' was used since it contains a comprehensive range of measures, as well as longitudinal follow up of individuals (University of Essex et al., 2018a). Understanding Society is a repeated annual UK longitudinal survey that measure demographic, social and health measures for roughly 40 000 individuals. At wave 2, a subset of participants took part in the Nurse Health Assessment survey that included additional health measures (University of Essex et al., 2018b). Participants were also invited to give a blood sample to measure biomarkers, as well as their genetic profile. The analytical sample used here are all individuals who took part in each part of data collection at wave 2 (n = 7243). Individuals who completed the Nurse Health Assessment in wave 3 were excluded due to their temporal mismatch. Table 1 presents sample characteristics.

A scoping exercise was first employed to select all variables within the survey where there was a plausible association to health. Variables were then assigned to a single domain (personal, social, health-related, biomarker and genetic). A conceptual framework was designed to capture core determinates of health by data type (see Appendix A). All relevant measures within each determinant were selected acknowledging that no single measure could best explain a concept by itself. Variables that were only asked to a small subset of participants or had a lot of missing data were excluded. A full list of all variables is presented in Appendix B.

Missing data is a common issue that few prediction models adequately address (B. A. Goldstein et al., 2017; Riley et al., 2016; Steyerberg et al., 2018). Complete case analysis was avoided due to the large number of variables with some form of missing data, as it would considerably reduce the sample size and may therefore introduce bias if missing data varies by socio-demographic characteristics. Imputation was opted against since it was hypothesised that the machine learning algorithms might be able to learn insights from missing data and that certain values add no value to the model. Missing data were modelled through recoding data as 0, with all variables coded to have positive values. Missing outcome data were excluded from analyses.

The outcome variable was "Whether an individual has a long-standing illness or disability" (yes (1) or no (0)). The variable is commonly used in health and social surveys to measure the presence of chronic health conditions due its ease of collection and has been shown to be a valid measure of actual health (Manor et al., 2001). Specific health conditions had low incidence during the study time period (e.g. hypertension was the most prevalent condition (22.7% at baseline) yet only 4.5% of individuals developed it five years post-baseline), meaning that they were less appropriate for inclusion.

Four models were analysed for predicting an individual's future health: (1) one-year post-baseline for all individuals; (2) five-years post-baseline for all individuals; (3) one-year post-baseline for only individuals at baseline with no long-standing illness; (4) five-years post-baseline for all individuals at baseline with no long-standing illness.

**2.2 Statistical analysis**

Three methods were applied. Firstly, artificial neural networks were selected due to their promise demonstrated elsewhere for predicting some medical outcomes (Beam & Kohane, 2018; Esteva et al., 2019; Gulshan et al., 2016). A simple feed forward model was used which iteratively learns what input values best predict an outcome through applying interconnected weights to input values (Efron & Hastie, 2016). A deep learning approach was tested through including additional layers to the model to account for potential complexity (i.e. modelling interacting relationships between combinations of weights), however this only resulted in marginal improvements in model fit (e.g. ~0.05 model accuracy improvement) for the first three additional layers before declining in performance. A parsimonious model, alternatively known as a shallow learning model, was selected that included only 3 layers – an input (values of individuals), intermediate (weights) and output (predictions of outcome) layer. This was kept consistent across all models to aid comparisons, with only the number of nodes varying depending on the number of inputs.

Second, a tree-based approach was selected to contrast the neural networks since the approach tends to work effectively with structured data (similar to survey data), unlike neural networks that often perform better with unstructured data (Efron & Hastie, 2016). Extreme Gradient Boosting (XGBoost) was used since it commonly outperforms other tree-based methods (Chen & Guestrin, 2016). The method works through building an ensemble classifier through combining multiple weak (shallow) classifiers to predict an outcome.

Thirdly, a logistic regression model was then used to assess how the additional complexity brought by the machine learning approaches compares to the performance of a simpler and more commonly used approach for clinical prediction modelling (Christodoulou et al., 2019).

**2.3 Analytical design**

Data were first split into train (75%) and test (25%) samples. The training sample was then standardised (mean 0, standard deviation 1) since all predictors had different ranges which will make model learning difficult and/or misleading. The test sample were applied the same standardisation to match the distribution of the training data and avoid any data leakage through standardising together (Efron & Hastie, 2016).

Feature reduction was then employed to drop variables not contributing much information to models to improve computational efficiency, as well as select a parsimonious model to minimise overfitting (Steyerberg et al., 2018). LASSO regression was used since it helps to deal with multicollinearity in predictors, as well as being computational efficient which was important for the genetic data (Efron & Hastie, 2016). 10-fold cross-validation was used to optimise the regularization parameter (selecting its minimum value to identify the optimal number of variables correlated to the outcome), with the process iterated 100 times to minimise stochastic elements and avoid a local solution.

A model for each of the three methods was then fit on the training data for each of the five domains separately, and then including all domains together. 4-fold cross-validation was used to internally evaluate and refine the model. Both machine learning approaches were selected to maximise the precision metric since our interest is in predicting presence of long-standing illness or disability. Model specification was selected to minimise overfitting through manual refinement of network structure and hyperparameters (e.g. number of layers, tree depth, metrics for assessing model learning). Models were then evaluated through predicting the outcomes of the unseen test data, with model performance evaluated against the actual outcomes. Evaluation statistics were selected to capture different aspects of model performance, as well as selecting those most commonly used in reporting clinical prediction models (Damen et al., 2016; Fahey et al., 2018; B. A. Goldstein et al., 2017; Sun et al., 2017). Specifically, (i) how representative predictions are (accuracy and kappa); (ii) how well models predict our outcome of interest – poor health (precision and sensitivity); (iii) how well they predict good health (specificity); (iv) an overall measure of discrimination (AUC – Area Under Curve); and (v) threshold-dependent measures (AUC, Kappa and TSS – True Skill Statistic).

Model interpretation is paramount when using machine learning, not just because of the black box nature of many algorithms but also for revealing insights about relationships. Feature importance was estimated for the neural network models by examining the correlation between the input values and their relationship to the predicted probabilities of the outcome for individuals. Variables that are negatively correlated were good predictors of 0s (i.e. good health), with positive correlation associated to 1s (i.e. poor health). Individual Conditional Expectation (ICE) plots and Partial Dependency plots (PDPs) were calculated to measure the marginal effect of predictors on the outcome (Efron & Hastie, 2016; A. Goldstein et al., 2013). Feature importance can be derived from the XGBoost model through identifying which variables were commonly used to split data during model development (Chen & Guestrin, 2016).

All analytical code can be found at https://github.com/markagreen/ukhls_biomedical_fellowship.

## 3 Results

Table 1 presents key sample characteristics of the analytical sample in comparison to all individuals who took part in the Nurse Health Assessment and the full survey. Participants included in the analyses were not representative of the overall survey or the separate Nurse Health Assessment wave suggesting that our data and therefore models are biased. The analytical sample were older on average, as well as containing a higher proportion of individuals who had a limiting long-term illness and who had no educational qualifications (although little difference was noted for the highest educational category).

**Table 1: Unweighted sample characteristics in comparison to Understanding Society.**

|  | Analytical Sample | Nurse Health Assessment | Full wave |
|---|---|---|---|
| Total sample size | 7 243* | 15 646 | 54 564 |
| Males (%) | 43.84 | 43.80 | 45.87 |
| Females (%) | 56.16 | 56.20 | 54.13 |
| Mean Age (standard deviation) | 52.20 (16.9) | 50.47 (17.8) | 46.63 (18.5) |
| No qualifications (%) | 26.51 | 14.91 | 15.99 |
| Secondary to Degree (%) | 52.33 | 62.68 | 62.94 |
| Degree or higher (%) | 21.16 | 22.41 | 21.07 |
| Limiting Long-Term Illness (%) | 39.75 | 39.52 | 34.36 |

* Full sample size excluding records with missing outcome records was 6830

Appendix B displays the overall results of the variable reduction analyses using LASSO regression. In sum, few biomarkers were dropped with the genetic data experiencing the largest proportion of variables dropped at this stage (partly reflecting a lack of variation in SNPs across the sample). Sex was dropped for three models suggesting little difference between males and females in short-term future risk. Fewer predictors were included when models only considered healthy individuals at baseline.

Table 2 presents results from model evaluation on the test data for each of the three methods considering all data at baseline. Results were similar for both 1-year and 5-year predictions. The health-related data were consistently the strongest set of predictors across the majority of model evaluation measures. The social measures and biomarkers were second best depending on which measure was selected (e.g. social data tended to perform better on sensitivity, with biomarkers better on specificity). Genetic data were consistently the poorest predictors; for example, their AUC suggested that they performed almost no different to random chance. The personal data results were largely driven by the inclusion of age. Indeed, re-running the models with only age as a predictor produced similar findings to the personal domain alone (e.g. Model 1 Neural Net AUC = 0.6392). Using all variables together only resulted in marginal improvements across most metrics when compared to the performance of the health-related data by itself.

The performance of neural nets was similar to the logistic regression model across most model fit statistics. XGBoost performed better on TSS, sensitivity and precision (but not specificity) than the other two approaches, suggesting it was better at predicting 1s (i.e. poor health). The neural net approach outperformed the other two approaches when considering AUC. No method performed consistent well across all evaluation metrics.

**Table 2: Comparing model fit for predicting the health status of participants one year from baseline**

| | Data Types | | | | | |
|---|---|---|---|---|---|---|
| | Personal | Social | Health-related | Biomarkers | Genetics | All |
| **Model 1: All individuals at baseline and whether they had a Limiting Long-Term Illness one year later (40.1% did); n = 6830** | | | | | | |
| Neural Net | | | | | | |
| Accuracy | 0.6163 | 0.6579 | 0.7077 | 0.6555 | 0.5483 | 0.6989 |
| Kappa | 0.1577 | 0.2572 | 0.3619 | 0.2398 | 0.0057 | 0.3514 |
| Sensitivity | 0.391 | 0.4737 | 0.5233 | 0.4271 | 0.2962 | 0.5474 |
| Specificity | 0.7601 | 0.7754 | 0.8253 | 0.8013 | 0.7092 | 0.7956 |
| Precision | 0.5098 | 0.5738 | 0.6566 | 0.5784 | 0.394 | 0.6308 |
| AUC | 0.6392 | 0.6735 | 0.7698 | 0.678 | 0.5042 | 0.7406 |
| TSS | 0.1511 | 0.2491 | 0.3486 | 0.2284 | 0.0054 | 0.343 |
| XGBoost | | | | | | |
| Accuracy | 0.6245 | 0.6596 | 0.7217 | 0.662 | 0.58 | 0.7317 |
| Kappa | 0.1643 | 0.246 | 0.3916 | 0.2473 | -0.0136 | 0.4129 |
| Sensitivity | 0.661 | 0.6876 | 0.7396 | 0.6864 | 0.6073 | 0.7462 |
| Specificity | 0.526 | 0.5879 | 0.6813 | 0.5957 | 0.3646 | 0.6987 |
| Precision | 0.7898 | 0.8109 | 0.8397 | 0.8215 | 0.8829 | 0.8493 |
| AUC | 0.5776 | 0.6167 | 0.6883 | 0.6168 | 0.4941 | 0.6983 |
| TSS | 0.187 | 0.2755 | 0.4209 | 0.2821 | -0.0281 | 0.4449 |
| Logistic Regression | | | | | | |
| Accuracy | 0.6134 | 0.6848 | 0.7042 | 0.6596 | 0.5718 | 0.7024 |
| Kappa | 0.1388 | 0.3028 | 0.3458 | 0.2408 | 0.0514 | 0.3557 |
| Sensitivity | 0.3449 | 0.4586 | 0.4842 | 0.4045 | 0.3098 | 0.5383 |
| Specificity | 0.7821 | 0.8292 | 0.8445 | 0.8225 | 0.739 | 0.8071 |
| Precision | 0.5054 | 0.6315 | 0.6653 | 0.5925 | 0.431 | 0.6404 |
| AUC | 0.5655 | 0.6439 | 0.6644 | 0.6135 | 0.5244 | 0.6727 |
| TSS | 0.127 | 0.2878 | 0.3287 | 0.227 | 0.0488 | 0.3454 |

**Model 2: All individuals at baseline and whether they had a Limiting Long-Term Illness five years later (42.7% did); n = 5220**

<u>Neural Net</u>

| | | | | | | |
|---|---|---|---|---|---|---|
| Accuracy | 0.6284 | 0.6613 | 0.6996 | 0.6521 | 0.5318 | 0.6851 |
| Kappa | 0.1978 | 0.287 | 0.3623 | 0.2607 | -0.0015 | 0.3437 |
| Sensitivity | 0.3821 | 0.5082 | 0.5247 | 0.4644 | 0.298 | 0.5722 |
| Specificity | 0.8061 | 0.7718 | 0.8259 | 0.7876 | 0.7005 | 0.7665 |
| Precision | 0.5871 | 0.6164 | 0.685 | 0.612 | 0.4179 | 0.6388 |
| AUC | 0.6494 | 0.7032 | 0.7477 | 0.6932 | 0.5115 | 0.7323 |
| TSS | 0.1882 | 0.28 | 0.3506 | 0.252 | -0.0015 | 0.3387 |

<u>XGBoost</u>

| | | | | | | |
|---|---|---|---|---|---|---|
| Accuracy | 0.6222 | 0.6674 | 0.7027 | 0.6467 | 0.5425 | 0.6973 |
| Kappa | 0.2119 | 0.297 | 0.3741 | 0.2487 | -0.018 | 0.3637 |
| Sensitivity | 0.6618 | 0.6862 | 0.7166 | 0.6663 | 0.5759 | 0.7138 |
| Specificity | 0.5556 | 0.6299 | 0.6763 | 0.6044 | 0.3975 | 0.6667 |
| Precision | 0.715 | 0.7876 | 0.8074 | 0.785 | 0.8061 | 0.7995 |
| AUC | 0.6043 | 0.6443 | 0.6825 | 0.6201 | 0.4917 | 0.6776 |
| TSS | 0.2174 | 0.3161 | 0.3929 | 0.2707 | -0.0266 | 0.3805 |

<u>Logistic Regression</u>

| | | | | | | |
|---|---|---|---|---|---|---|
| Accuracy | 0.6261 | 0.6705 | 0.6874 | 0.6475 | 0.5471 | 0.6897 |
| Kappa | 0.2074 | 0.3064 | 0.3394 | 0.2494 | 0.0379 | 0.3546 |
| Sensitivity | 0.4424 | 0.5192 | 0.5265 | 0.4516 | 0.34 | 0.585 |
| Specificity | 0.7586 | 0.7797 | 0.8034 | 0.7889 | 0.6966 | 0.7652 |
| Precision | 0.5694 | 0.6297 | 0.659 | 0.6069 | 0.4471 | 0.6426 |
| AUC | 0.6005 | 0.6494 | 0.665 | 0.6202 | 0.5183 | 0.6751 |
| TSS | 0.201 | 0.2989 | 0.3299 | 0.2405 | 0.0366 | 0.3502 |

Note: AUC = Area Under Curve, TSS = True Skill Statistic.

**Table 3: Comparing model fit for predicting the health status of participants five years from baseline**

| | Data Types | | | | | |
|---|---|---|---|---|---|---|
| | Personal | Social | Health-related | Biomarkers | Genetics | All |

**Model 3: All individuals with no Limiting Long-Term Illness at baseline and whether they reported the outcome one year later (16.4% did); n = 4101**

<u>Neural Net</u>

| | | | | | | |
|---|---|---|---|---|---|---|
| Accuracy | 0.8341 | 0.8293 | 0.8332 | 0.8351 | 0.8215 | 0.8254 |
| Kappa | 0 | 0.0133 | 0.043 | 0.0175 | 0.0128 | 0.087 |
| Sensitivity | 0 | 0.0176 | 0.0353 | 0.0118 | 0.0294 | 0.0882 |
| Specificity | 1 | 0.9906 | 0.9918 | 0.9988 | 0.9789 | 0.9719 |
| Precision | N/A | 0.2727 | 0.4615 | 0.6667 | 0.2174 | 0.3846 |
| AUC | 0.6036 | 0.5773 | 0.6006 | 0.5836 | 0.4835 | 0.5896 |
| TSS | 0 | 0.0082 | 0.0271 | 0.0106 | 0.0083 | 0.0601 |
| XGBoost | | | | | | |
| Accuracy | 0.8293 | 0.8312 | 0.8244 | 0.8302 | 0.8312 | 0.8283 |
| Kappa | 0.0133 | 0.002 | -0.0036 | 0.0152 | 0.002 | 0.0187 |
| Sensitivity | 0.8353 | 0.8343 | 0.8338 | 0.8355 | 0.8343 | 0.8358 |
| Specificity | 0.2727 | 0.2 | 0.1429 | 0.3 | 0.2 | 0.2857 |
| Precision | 0.9906 | 0.9953 | 0.986 | 0.9918 | 0.9953 | 0.9883 |
| AUC | 0.5041 | 0.5006 | 0.4989 | 0.5047 | 0.5006 | 0.5059 |
| TSS | 0.108 | 0.0343 | -0.0233 | 0.1355 | 0.0343 | 0.1215 |
| Logistic Regression | | | | | | |
| Accuracy | 0.8341 | 0.8332 | 0.8273 | 0.8361 | 0.8176 | 0.8117 |
| Kappa | 0 | 0.0058 | 0.0094 | 0.0271 | -0.0088 | 0.0665 |
| Sensitivity | 0 | 0.0059 | 0.0176 | 0.0176 | 0.0176 | 0.0941 |
| Specificity | 1 | 0.9977 | 0.9883 | 0.9988 | 0.9766 | 0.9544 |
| Precision | N/A | 0.3333 | 0.2308 | 0.75 | 0.1304 | 0.2909 |
| AUC | 0.5 | 0.5018 | 0.503 | 0.5082 | 0.4971 | 0.5243 |
| TSS | 0 | 0.0036 | 0.0059 | 0.0164 | -0.0058 | 0.0485 |

**Model 4: All individuals with no Limiting Long-Term Illness at baseline and whether they reported the outcome five years later (24.3% did); n = 3126**

| | | | | | | |
|---|---|---|---|---|---|---|
| Neural Net | | | | | | |
| Accuracy | 0.7516 | 0.749 | 0.7439 | 0.7593 | 0.7286 | 0.7529 |
| Kappa | 0 | 0.0716 | 0.0571 | 0.0695 | 0.0415 | 0.1951 |
| Sensitivity | 0 | 0.0825 | 0.0773 | 0.0567 | 0.0928 | 0.2371 |
| Specificity | 1 | 0.9693 | 0.9642 | 0.9915 | 0.9387 | 0.9233 |
| Precision | N/A | 0.4706 | 0.4167 | 0.6875 | 0.3333 | 0.5055 |
| AUC | 0.5412 | 0.6129 | 0.6608 | 0.6228 | 0.5322 | 0.6209 |
| TSS | 0 | 0.0518 | 0.0415 | 0.0482 | 0.0315 | 0.1604 |
| XGBoost | | | | | | |
| Accuracy | 0.7452 | 0.7465 | 0.749 | 0.7478 | 0.7452 | 0.7542 |
| Kappa | 0.0685 | 0.0297 | 0.1098 | 0.0691 | 0.0126 | 0.1199 |
| Sensitivity | 0.7615 | 0.7556 | 0.7682 | 0.7614 | 0.7533 | 0.7694 |
| Specificity | 0.4359 | 0.4 | 0.4808 | 0.4571 | 0.3333 | 0.5208 |
| Precision | 0.9625 | 0.9796 | 0.954 | 0.9676 | 0.983 | 0.9608 |
| AUC | 0.5251 | 0.5104 | 0.5414 | 0.5251 | 0.5044 | 0.5448 |
| TSS | 0.1974 | 0.1556 | 0.249 | 0.2185 | 0.0866 | 0.2902 |
| Logistic Regression | | | | | | |
| Accuracy | 0.7567 | 0.7465 | 0.7516 | 0.758 | 0.7337 | 0.735 |
| Kappa | 0.0356 | 0.0621 | 0.1229 | 0.0669 | 0.0596 | 0.1648 |
| Sensitivity | 0.0258 | 0.0773 | 0.1392 | 0.0567 | 0.1031 | 0.2423 |
| Specificity | 0.9983 | 0.9676 | 0.954 | 0.9898 | 0.9421 | 0.8978 |
| Precision | 0.8333 | 0.4412 | 0.5 | 0.6471 | 0.3704 | 0.4393 |
| AUC | 0.512 | 0.5225 | 0.5466 | 0.5232 | 0.5226 | 0.57 |
| TSS | 0.0241 | 0.0449 | 0.0932 | 0.0465 | 0.0452 | 0.1401 |

Note: AUC = Area Under Curve, TSS = True Skill Statistic.

When data were restricted to only individuals with no limiting long-term illness at baseline (Table 3),

model accuracy was consistently improved for all data types. The improvement in model accuracy did not reflect a more insightful model though. The neural net and logistic regression models were poor at predicting 1s (i.e. low sensitivity values), mainly predicting 0s (i.e. large specificity values). The low incidence of the outcome variable meant that through models predicting good health for most individuals, it resulted in a high model accuracy. The XGBoost models performed better when comparing precision, sensitivity and TSS statistics, although performed poorly on specificity compared to the other models. The neural network outperforms the other methods on AUC. The mixed performance across the model fit statistics suggest no single method was the best.

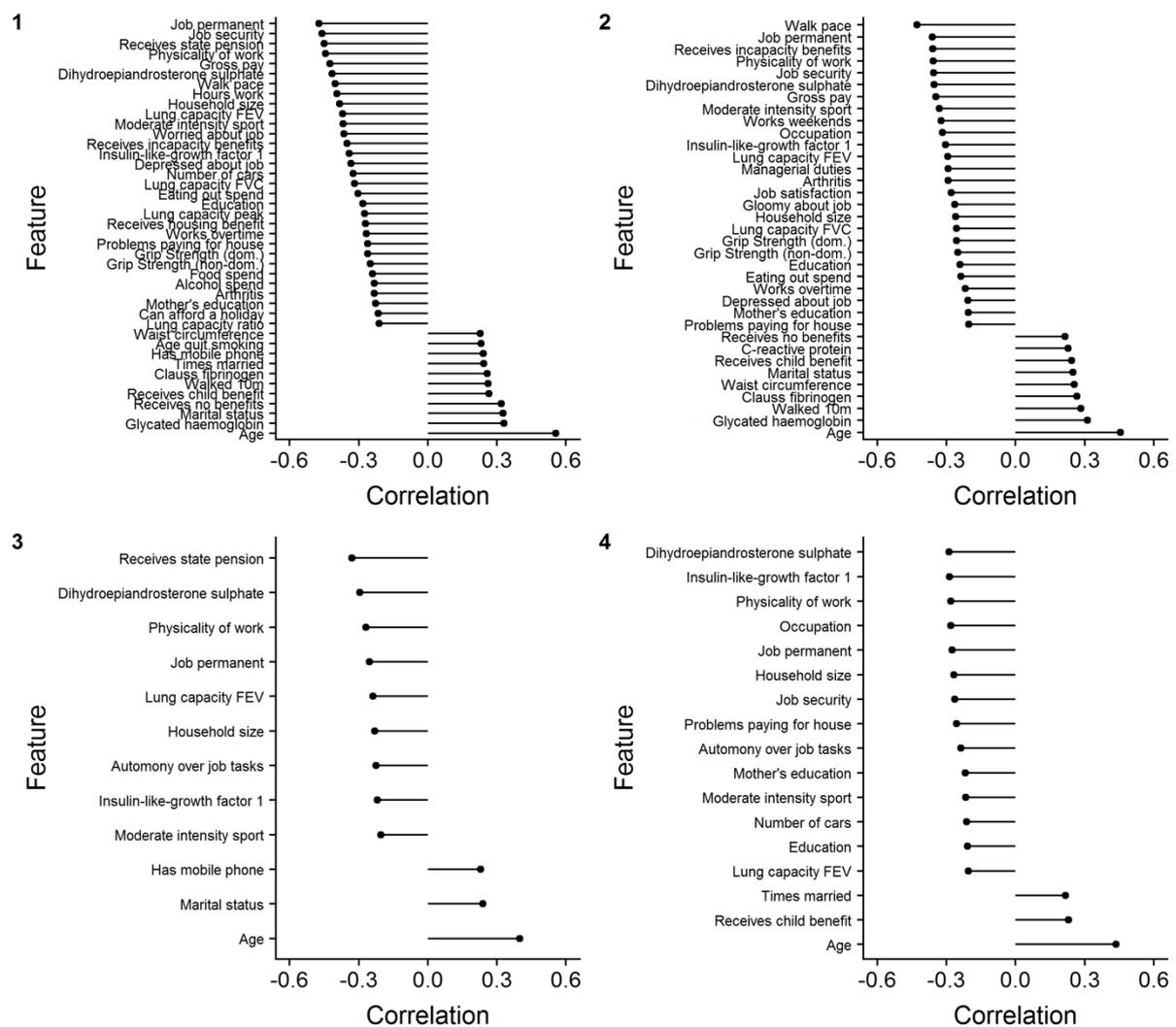

Figure 1: Correlation of input values to overall predicted outcome values for each neural network (models numbered by facet).

Feature importance for the neural net models with all variables is displayed in Figure 1. There were more predictors of good health (negative correlations) than compared to poor health (positive correlations) across all models, as well as in the models with that included all individuals. Age stood out as the strongest predictor of ill health (1s), with a mixture of social (e.g. job permanent, job security), health-related (e.g. walk pace, physicality of work) and biomarkers (e.g. dehydroepiandrosterone sulphate, insulin-like-growth factor 1) as stronger predictors of good health (0s). There were no SNPs in any of the plots. Figures C1-C4 (Appendix) present plots of model interpretation for the best predictors in each model. Relationships are mostly linear (partly due to the low number of layers that may capture non-linear relationships) and largely followed expected directions.

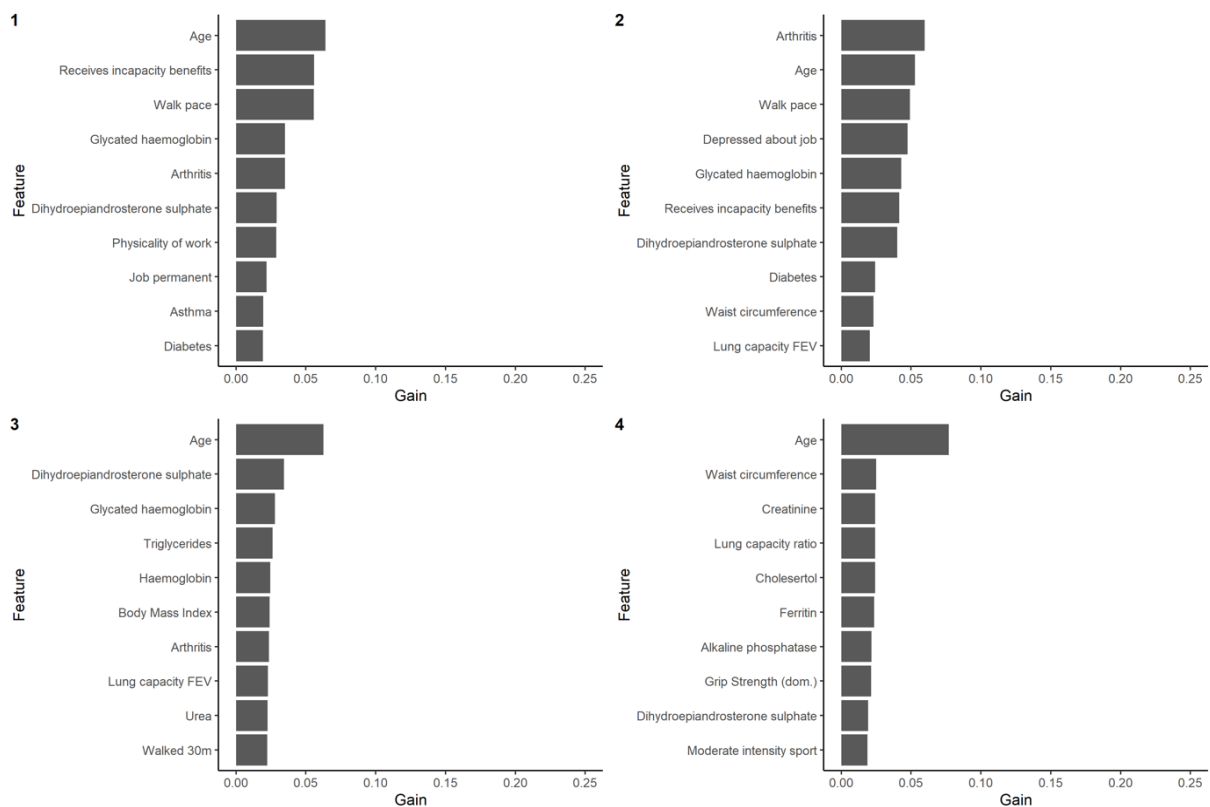

**Figure 2: Feature importance for top 10 performing predictors by gain metric in each XGBoost model (models numbered by facet).**

Figure 2 presents the feature importance for the XGBoost models (Figure 2). While age was also important in each model, there are some differences. Biomarkers were more prominent across all models, with fewer social measures included. No SNPs were found to be important again. Figures D1-D4 (Appendix) present plots for interpretation of predictor's relationships to outcomes in each model.

**4 Discussion**

**4.1 Key Results**

This paper demonstrates one of the most comprehensive applications of machine learning to evaluate the predictive ability of a diverse range of data types for predicting future risk of ill health. Health-related and wellbeing measures were the strongest predictors of future health status, with genetic data the poorest performing. No single methodology performed consistently the best across our evaluation statistics and both machine learning approaches largely performed similar to logistic regression models. Neural nets performed better on AUC and XGBoost best on sensitivity and precision metrics. Feature importance revealed the best predictors to be age, as well as a mixture of social (e.g. job permanent, job security), health-related (e.g. walk pace, physicality of work) and biomarkers (e.g. dehydroepiandrosterone sulphate, insulin-like-growth factor 1).

**4.2 Interpretation**

Health-related data performed the best at predicting future ill health across the models, although no clear single variable stood out as the key measure. Social measures performed just as well as the biomarker data reflecting the importance of the social determinants of health (Green et al., 2017; Merlo, 2014; Taylor-Robinson & Kee, 2019). The poor performance of the genetic data may reflect either their poor discriminatory power in the predicting ill health, or the lack of variation within the sample due to the small sample size.

With genetic and biomarker data not dominating model performance, it suggests that traditional and less intrusive data collection methods can be just as important than these newer forms of data for predicting health (Taylor-Robinson & Kee, 2019). Age by itself performed fairly well and was an important feature across all models. Similar findings have been observed across other predictive models; for example, the strongest predictive factor in the Framingham Risk score is age (Riley et al., 2016). The results suggest that we might not need the added complexity of hundreds to thousands of variables, prioritising only those which matter most. However, the inclusion of additional variables still improved the value of the model (e.g. model 1 neural network AUC for just age alone was 0.60, but 0.74 for all variables). This is important since increasing model fit becomes harder to achieve as a model continues to improve.

How well do the models perform? Reviews of the literature for a range of outcomes and models find AUCs commonly between 0.7 and 0.8, although the range often includes models as low as 0.6 (Fahey et al., 2018; Siontis et al., 2012; Sun et al., 2017). This suggests that some models in this study perform well, with others at the lower end of performance. Poor model performance may reflect unsuitable methods or study design, insufficient measures and variables available in Understanding Society, or even the folly of trying to accurately predict limiting long-term illness. Model evaluation statistics for the simpler logistic regression models were largely similar or not substantially lower than the machine learning methods. The lack of any substantial improvement of machine learning models compared to logistic regression has been noted elsewhere (Christodoulou et al., 2019). However, this finding was not consistent across all metrics and models. Observed AUCs were generally higher for the neural networks, with XGBoost performing better on those metrics measuring how well it predicted poor health (e.g. precision or sensitivity).

The lack of a conisderable improvement in the machine learning methods may reflect data issues. While both machine learning approaches can be used effectively with small data (e.g. (Beam, 2017)), machine learning algorithms are 'data hungry' requiring large quantities of information to train models particularly for detecting non-linear relationships (Christodoulou et al., 2019; Efron & Hastie, 2016). Relationships may not be complex (or non-linear), or the outcome variables might be weakly associated to our predictors. The outcome variable is general in concept and includes a range of health conditions with diverse causes, which may obscure prediction. Quality of data is more important than quantity of data, particularly to extract signal from noise.

An interpretation that the traditional logistic regression model is just as effective as machine learning approaches is misleading. The logistic regression models did not frequently outperform the machine learning methods when evaluating the model fit statistics. The performance of the logistic regression models is also inflated due to the incorporation of the LASSO regression to reduce the number of predictors. Without this design step, the use of these models becomes problematic and cannot always be used (especially with the genetic data). Indeed, the logistic regression models struggle to fit the models with all predictors included. The flexibility and scalability of machine learning approaches represent an important advantage here where we want to incorporate large quantities of data (Esteva et al., 2019).

The results suggest that even with the most comprehensive information about individuals and complex models to interpret it, we are still a long way off perfect discrimination between cases and non-cases. Perfectly predicting future risk of ill health is unlikely to be feasible (Green et al., 2017). However, this does not mean that such approaches are not useful. The four models presented may offer additional 'digital health' tools to supplement decision making, assist diagnoses and 'augment clinicians' (Saria et al., 2018). With models scalable and adaptable, they can be easily deployed to increasingly complex health systems and massive quantities of existing electronic health and administrative records (Esteva et al., 2019). Reinforcing these methods with approaches that build and demonstrate causal relationships in their decision making will further help to improve their validity.

**4.3 Limitations**

While the overall UKHLS sample is representative of the UK, the sample used in our analyses exhibits selection bias (Table 1) and the cohort suffers from attrition (Lynn & Borkowska, 2018). Both of these issues are associated to our outcome variable, as well as key predictors. This limits the generalisability of the findings which is a common issue in developing effective predictive models (Fahey et al., 2018). Greater investigation into the impact of missing data and comparisons to other approaches was also required, including whether the methods learnt to detect missing observations.

The outcome variable used could have been more specific such as a health outcome or response to an intervention to improve the usefulness of the model. This was a limitation in the low levels of incidence of recorded diseases and repeating the analyses in a larger dataset may have helped.

The baseline predictors are cross-sectional and this snapshot may poorly reflect an individual's circumstances. While the UKHLS is longitudinal, not all predictors were collected over time limiting our ability to extend the model longitudinally. Recurrent learning and alternative models that can incorporate sequencing of data may offer potential here for improving predictive model performance.

It is important to externally validate the models with an independent dataset (Christodoulou et al., 2019; Damen et al., 2016). There may be some indirect spill over of information even though the data were split into train and test samples due to the nature of data collection. External validation was difficult here due to the number of variables included; there were no alternative datasets available that contained the range of predictors. The success of simpler approaches such as the Framingham Risk Score has been their deployability in health care systems and different contexts (B.

A. Goldstein et al., 2017; Siontis et al., 2012), which may hinder the results from this study and machine learning applications in general.

## 5 Conclusions

The growing interest in incorporating complexity into health-related enquiry may not be the panacea many advocates argue. This study has demonstrated one example where utilising more complex data and methods did not translate to better models or understanding of the determinants of ill health. It is clear that even with all the added complexity possible, we are still a long way off achieving any level of personalised medicine (Taylor-Robinson & Kee, 2019). Focusing on the value offered through the careful consideration of applications for new data and methods is key for advancing the evidence base. For example, machine learning is merely an extension of statistics; it isn't a tool that can generate meaningful results in every situation (Beam & Kohane, 2018).

**Acknowledgements**

This work was supported by the Understanding Society Biomedical Data Fellowship Programme (a subcontract of ESRC grant ES/M008592/1). I would like to thank the advice and mentorship offered by the Understanding Society team particularly Paul Clarke, Meena Kumari, Micheala Benzeval, Steve Pudney and Yanchun Bao, as well as from all of the other Fellows. MAG was responsible for all parts of completing the study including data analysis and writing. MAG had full access to the data and was responsible for submission of the paper. Access to the genetic data was approved by the METADAC committee on 11[th] May 2018 (ref: MDAC-2018-0008-02A-GREEN). The datasets supporting the conclusions of this article are available in the UK Data Archive repository, available at http://doi.org/10.5255/UKDA-SN-6614-12 and http://doi.org/10.5255/UKDA-SN-7251-3. All analytical code: https://github.com/markagreen/ukhls_biomedical_fellowship. No conflicts of interest are declared.

**Appendices**

For the paper: *Evaluating the performance of personal, social, health-related, biomarker and genetic data for predicting an individual's future health using machine learning*: A longitudinal analysis

Contents:

Note:

Four models were analysed for predicting an individual's future health: (1) one-year post-baseline for all individuals; (2) five-years post-baseline for all individuals; (3) one-year post-baseline for only individuals at baseline with no long-standing illness; (4) five-years post-baseline for all individuals at baseline with no long-standing illness.

**Appendix A – Conceptual framework for variable inclusion**

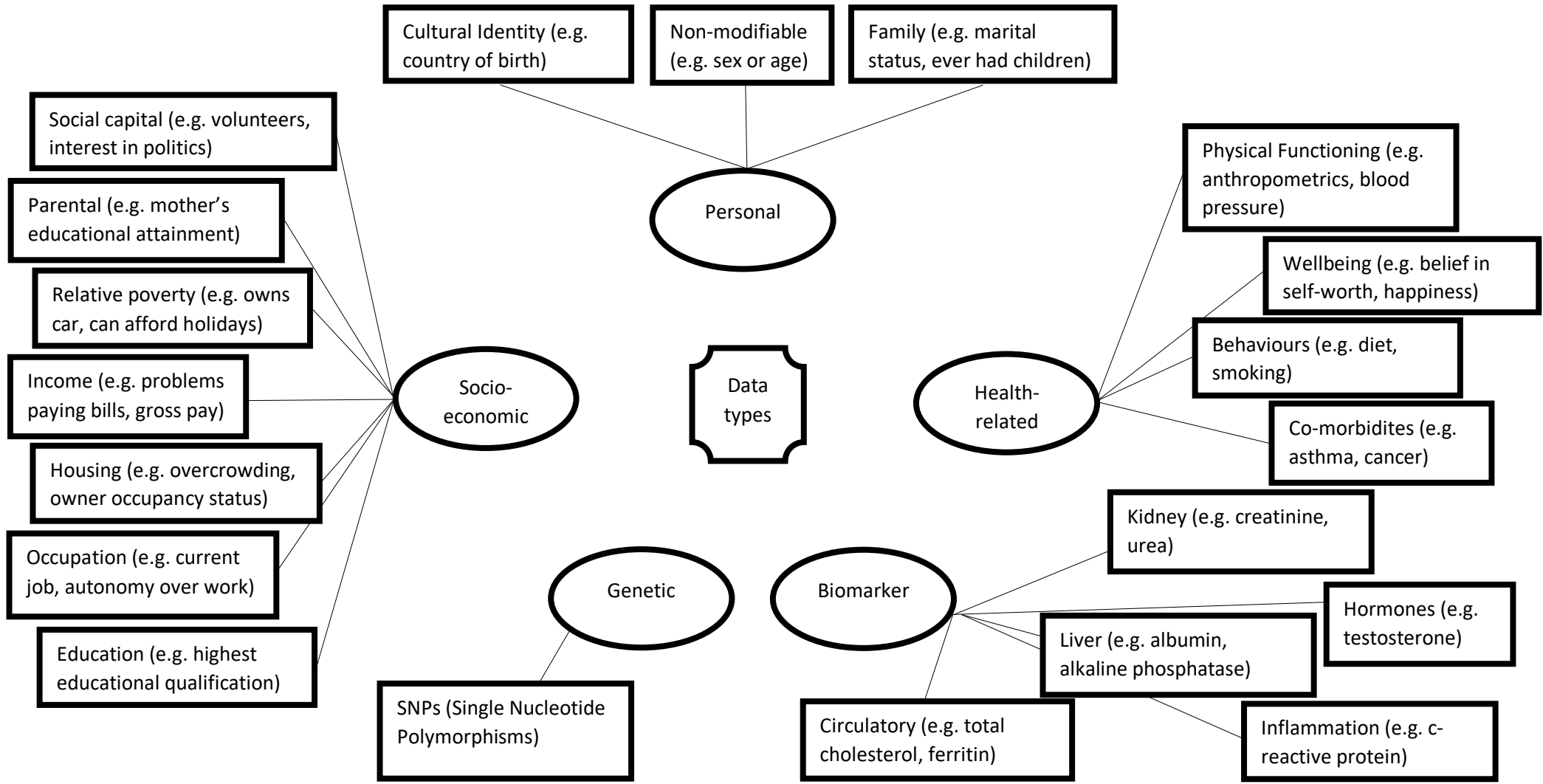

# Appendix B – Variables included in the analyses

The section details the variables included in the analyses including summary statistics for the variable reduction process (Table B1) and the specific variables included (Table B2). While not all variables selected from the framework scoping process were provided by the UK Data Archive due to disclosure issues, the excluded variables were minor in capturing each domain.

**Table B1: Summary of the LASSO regression results for variable selection by model.**

| Domain | Total variables | Variables selected | | | |
|---|---|---|---|---|---|
| | | Model 1 | Model 2 | Model 3 | Model 4 |
| Personal | 8 | 3 | 4 | 3 | 4 |
| Socioeconomic | 73 | 39 | 41 | 19 | 42 |
| Health-related | 55 | 49 | 45 | 30 | 43 |
| Biomarker | 20 | 18 | 16 | 13 | 17 |
| Genetics | 504,218 | 83 | 42 | 40 | 32 |
| Total | 504,374 | 192 | 148 | 105 | 138 |

**Table B2: Variable inclusion following LASSO regression analysis. Note: x means variable was selected.**

| Domain | Sub-Domain | Variable | Model 1 | Model 2 | Model 3 | Model 4 |
|---|---|---|---|---|---|---|

| Category | Subcategory | Variable | | | | |
|---|---|---|---|---|---|---|
| personal | non-modifiable | Sex | | x | | |
| personal | non-modifiable | Age | x | x | x | x |
| personal | family | Legal marital status | x | x | x | |
| personal | family | Number of times respondent married | x | | x | x |
| personal | family | Has any children responsible for | | | | |
| personal | identity | Born in UK | | | | |
| personal | identity | Country father born in | | | | |
| personal | identity | Country mother born in | | x | | x |
| socioeconomic | parental | Father's educational qualifications | | | | |
| socioeconomic | parental | Mother's educational qualifications | x | x | x | x |
| socioeconomic | parental | Father not working when respondent was 14 | x | | x | x |
| socioeconomic | parental | Father's occupation (SOC 2000) when respondent was 14 | x | x | | |
| socioeconomic | parental | Mother not working when respondent was 14 | | x | | x |
| socioeconomic | parental | Mother's occupation (SOC 2000) when respondent was 14 | x | | | |
| socioeconomic | education | Highest qualification | x | x | x | x |
| socioeconomic | housing | Number of people in household | x | x | x | x |
| socioeconomic | housing | Number of bedrooms | | | x | x |
| socioeconomic | housing | Number of beds | | | | x |
| socioeconomic | housing | House owned or rented | | | x | x |
| socioeconomic | housing | Value of property: home owners | x | x | | x |
| socioeconomic | housing | Household has central heating | | | | |
| socioeconomic | income | Problems paying for housing | x | x | | x |
| socioeconomic | income | Problems paying for council tax | | | | x |
| socioeconomic | income | Do you receive: Unemployment benefits | | | | |
| socioeconomic | income | Do you receive: Disability benefits | x | x | x | x |
| socioeconomic | income | Do you receive: Pensions | x | x | x | x |
| socioeconomic | income | Do you receive: Receives SERPS | x | x | x | |
| socioeconomic | income | Do you receive: Child Benefit | x | | x | x |
| socioeconomic | income | Do you receive: Child Tax Credit | x | x | | x |
| socioeconomic | income | Do you receive: Family benefits | x | x | | |
| socioeconomic | income | Do you receive: Tax Credits | x | | | x |
| socioeconomic | income | Do you receive: Housing related benefits | x | | x | |
| socioeconomic | income | Do you receive: Other sources | x | | | x |
| socioeconomic | income | Do you receive: No benefit received | x | x | | x |
| socioeconomic | income | Amount received in interest/dividends | x | | | x |
| socioeconomic | income | Subjective financial situation - current | x | x | x | x |
| socioeconomic | income | Whether contributes to personal pension | | x | | |
| socioeconomic | income | Whether saves | | x | | x |
| socioeconomic | income | Monthly amount saved | | | | |
| socioeconomic | income | Gross pay at last payment | x | x | | |
| socioeconomic | social capital | Volunteer in last 12 months | x | x | | x |
| socioeconomic | social capital | Donated money to charity | x | x | x | x |
| socioeconomic | social capital | Supports a particular political party | x | x | | x |
| socioeconomic | social capital | Party would vote for tomorrow | | x | | x |
| socioeconomic | social capital | which political party closest to | x | x | | |
| socioeconomic | social capital | Level of interest in politics | x | | x | x |
| socioeconomic | relative deprivation | Material deprivation: holiday | x | x | | x |
| socioeconomic | relative deprivation | Material deprivation: social meal/drink | x | x | x | x |
| socioeconomic | relative deprivation | Material deprivation: shoes | | | | |
| socioeconomic | relative deprivation | Material deprivation: house | x | x | | x |
| socioeconomic | relative deprivation | Material deprivation: contents insurance | | x | | x |
| socioeconomic | relative deprivation | Material deprivation: savings | x | | x | |
| socioeconomic | relative deprivation | Material deprivation: furniture | | | | x |
| socioeconomic | relative deprivation | Material deprivation: electrical goods | x | | | |
| socioeconomic | relative deprivation | Do you have a PC? | | | | |
| socioeconomic | relative deprivation | Has access to the internet from home | | | | x |
| socioeconomic | relative deprivation | Number of cars in household | x | | | x |
| socioeconomic | relative deprivation | Has use of car or van | | x | | |
| socioeconomic | relative deprivation | Has mobile phone | x | x | x | x |
| socioeconomic | relative deprivation | Has use of tumble dryer | x | x | | x |

| | | | | | | | |
|---|---|---|---|---|---|---|---|
| socioeconomic | relative deprivation | Has use of sky or cable | | x | | x | x |
| socioeconomic | relative deprivation | Has use of dishwasher | | | | | x |
| socioeconomic | relative deprivation | Has use of a cd player | | | | | |
| socioeconomic | occupation | Current economic activity | | | | | |
| socioeconomic | occupation | Current job: permanent or temporary | | x | x | x | x |
| socioeconomic | occupation | Occupation (SOC2000): current job | | | x | | x |
| socioeconomic | occupation | Managerial duties current job check | | | x | | |
| socioeconomic | occupation | Number of hours normally employed per week | x | | | | |
| socioeconomic | occupation | Number of overtime hours in normal week | x | x | | | |
| socioeconomic | occupation | Job satisfaction | | | x | | |
| socioeconomic | occupation | Usually works weekends | | | x | | |
| socioeconomic | occupation | Autonomy over job tasks | | | | x | x |
| socioeconomic | occupation | Autonomy over work pace | | | | | |
| socioeconomic | occupation | Autonomy over work manner | | | | | |
| socioeconomic | occupation | Autonomy over task order | | | | | |
| socioeconomic | occupation | Autonomy over work hours | | | | | |
| socioeconomic | occupation | Feels tense about job | | | | | |
| socioeconomic | occupation | Feels uneasy about job | | | | | |
| socioeconomic | occupation | Feels worried about job | x | | | | |
| socioeconomic | occupation | Feels depressed about job | x | x | | | x |

| Category | Subcategory | Variable | | | | |
|---|---|---|---|---|---|---|
| socioeconomic | occupation | Feels gloomy about job | | x | | |
| socioeconomic | occupation | Feels miserable about job | | | | |
| socioeconomic | occupation | Job security in next 2 months | x | x | | x |
| socioeconomic | occupation | Has a second job | | x | x | x |
| health-related | Physical functioning | Standing height (cm) for reliable measurements | x | x | x | x |
| health-related | Physical functioning | Standing weight (kg) for reliable measurements | x | x | x | x |
| health-related | Physical functioning | Body fat percentage for valid measurements | x | x | | |
| health-related | Physical functioning | Waist circumference valid (cm) | x | x | x | x |
| health-related | Physical functioning | BMI includes valid measurements | x | x | x | x |
| health-related | Physical functioning | Valid max grip strength dominant hand (kg) | x | x | | x |
| health-related | Physical functioning | Valid max grip strength non-dominant hand (kg) | x | x | | |
| health-related | Physical functioning | Forced vital capacity (total air blown out, litres) Highest | x | x | | x |
| health-related | Physical functioning | Forced expiratory volume in 1 second (amount of air blown out in 1 sec in litres) highest | x | x | x | x |
| health-related | Physical functioning | Ratio of FEV1/FVC (based on highest) | x | x | | x |
| health-related | Physical functioning | Peak expiratory flow (speed of air moving out of lungs, litres per sec) highest values | | | | |
| health-related | Physical functioning | Mean systolic blood pressure (valid) | | | | |
| health-related | Physical functioning | Mean diastolic blood pressure (valid) | x | x | | |
| health-related | Physical functioning | Mean pulse (valid) | x | | | x |
| health-related | wellbeing | GHQ:Concentration | x | x | x | x |
| health-related | wellbeing | GHQ:Loss of sleep | x | x | x | x |
| health-related | wellbeing | GHQ:Playing a useful role | x | x | x | x |
| health-related | wellbeing | GHQ:Capable of making decisions | x | x | x | x |
| health-related | wellbeing | GHQ:Constantly under strain | x | x | x | x |
| health-related | wellbeing | GHQ:Problem overcoming difficulties | x | x | | x |
| health-related | wellbeing | GHQ:Enjoy day-to-day activities | x | x | x | x |
| health-related | wellbeing | GHQ:Ability to face problems | x | x | | x |
| health-related | wellbeing | GHQ:Unhappy or depressed | x | x | | x |
| health-related | wellbeing | GHQ:Losing confidence | x | x | x | x |
| health-related | wellbeing | GHQ:Believe in self worth | x | x | x | x |
| health-related | wellbeing | GHQ:General happiness | x | x | | x |
| health-related | behaviours | Amount spent on food from supermarket | x | x | x | |
| health-related | behaviours | Amount spent on meals/snacks outside the home | x | x | x | x |
| health-related | behaviours | Amount spent on alcohol | x | x | x | x |
| health-related | behaviours | Usual type of dairy consumption | x | x | | x |
| health-related | behaviours | Type of bread eats most frequently | x | | | x |
| health-related | behaviours | Days each week eat fruit | x | x | x | x |
| health-related | behaviours | Days each week eat vegetables | x | x | x | |
| health-related | behaviours | Usual portion of fruit/veg eaten | x | x | x | |
| health-related | behaviours | Done walking at least 10 minutes | | | | x |
| health-related | behaviours | Number of days walked at least 10 minutes | x | x | | |
| health-related | behaviours | Number of days walked at least 30 minutes | x | x | x | x |
| health-related | behaviours | Average pace of walking | x | x | x | x |
| health-related | behaviours | Ever smoked cigarettes | | | | x |
| health-related | behaviours | Smoke cigarettes now | x | | | x |
| health-related | behaviours | Usual no. of cigarettes smoked per day | x | x | x | x |
| health-related | behaviours | Ever smoked cigarettes regularly | x | | | |
| health-related | behaviours | Age when last stopped smoking | x | x | x | x |
| health-related | behaviours | Physicality of job | x | x | x | x |
| health-related | behaviours | Moderate intensity sports frequency | x | x | x | x |

| category | subcategory | variable | | | | |
|---|---|---|---|---|---|---|
| health-related | behaviours | Mild intensity sports frequency | x | x | x | x |
| health-related | health conditions | Has been diagnosed with: Asthma | x | | | x |
| health-related | health conditions | Has been diagnosed with: Arthritis | x | x | x | x |
| health-related | health conditions | Has been diagnosed with: Cancer | x | x | x | x |
| health-related | health conditions | Has been diagnosed with: Clinical depression | x | | | x |
| health-related | health conditions | Has been diagnosed with: Diabetes | x | x | x | x |
| health-related | health conditions | Has been diagnosed with: Heart-related conditions | x | x | x | |
| health-related | health conditions | Has been diagnosed with: Hypertension | x | x | x | x |
| health-related | health conditions | Has been diagnosed with: Other | x | x | x | |
| biomarker | circulatory | Cholesterol (total) | x | x | x | x |
| biomarker | circulatory | HDL cholesterol | x | x | x | x |
| biomarker | circulatory | Triglycerides | x | x | x | x |
| biomarker | circulatory | Glycated haemoglobin | x | x | x | x |
| biomarker | inflammation | C-reactive protein | x | x | | x |
| biomarker | inflammation | Cytomegalovirus IgG | x | x | | x |
| biomarker | inflammation | Cytomegalovirus IgM | x | x | | x |
| biomarker | inflammation | Clauss fibrinogen | x | x | x | x |
| biomarker | circulatory | Haemoglobin | x | x | x | x |
| biomarker | circulatory | Ferritin | x | | | x |
| biomarker | liver | Albumin | x | x | | |
| biomarker | liver | Alkaline phosphatase | x | x | x | x |
| biomarker | liver | Alanine transaminase | x | x | | x |
| biomarker | liver | Aspartate transaminase | x | | x | |
| biomarker | liver | Gamma glutamyl transferase | x | x | x | |
| biomarker | kidney | Creatinine | x | x | x | x |
| biomarker | kidney | Urea | | | x | x |
| biomarker | hormones | Testosterone | | | | x |
| biomarker | hormones | Insulin-like-growth factor 1 | x | x | x | x |
| biomarker | hormones | Dihydroepiandrosterone sulphate | x | x | x | x |

# Appendix C – Neural network model interpretation

Individual Conditional Expectation (ICE) plots (Figures C1-C4) were used to estimate the marginal effect of the two predictors with the strongest correlation values from Figure 1 for both positive and negative associations. The plots allowed the investigation of model interpretability to identify how predictions were arrived at. In each plot, the yellow highlighted line represents the mean estimate (equivalent to partial dependency plot). The grey lines representing the association for a single observation (a random sample of 5% of the data are plotted), with points representing the measured value. The plots provide a measure of the uncertainty in associations and display the diversity of predicted relationships. Estimates have been left centred to aid interpretation of results (for each plot this represents missing data). Standardised values of inputs are plotted to match model inputs.

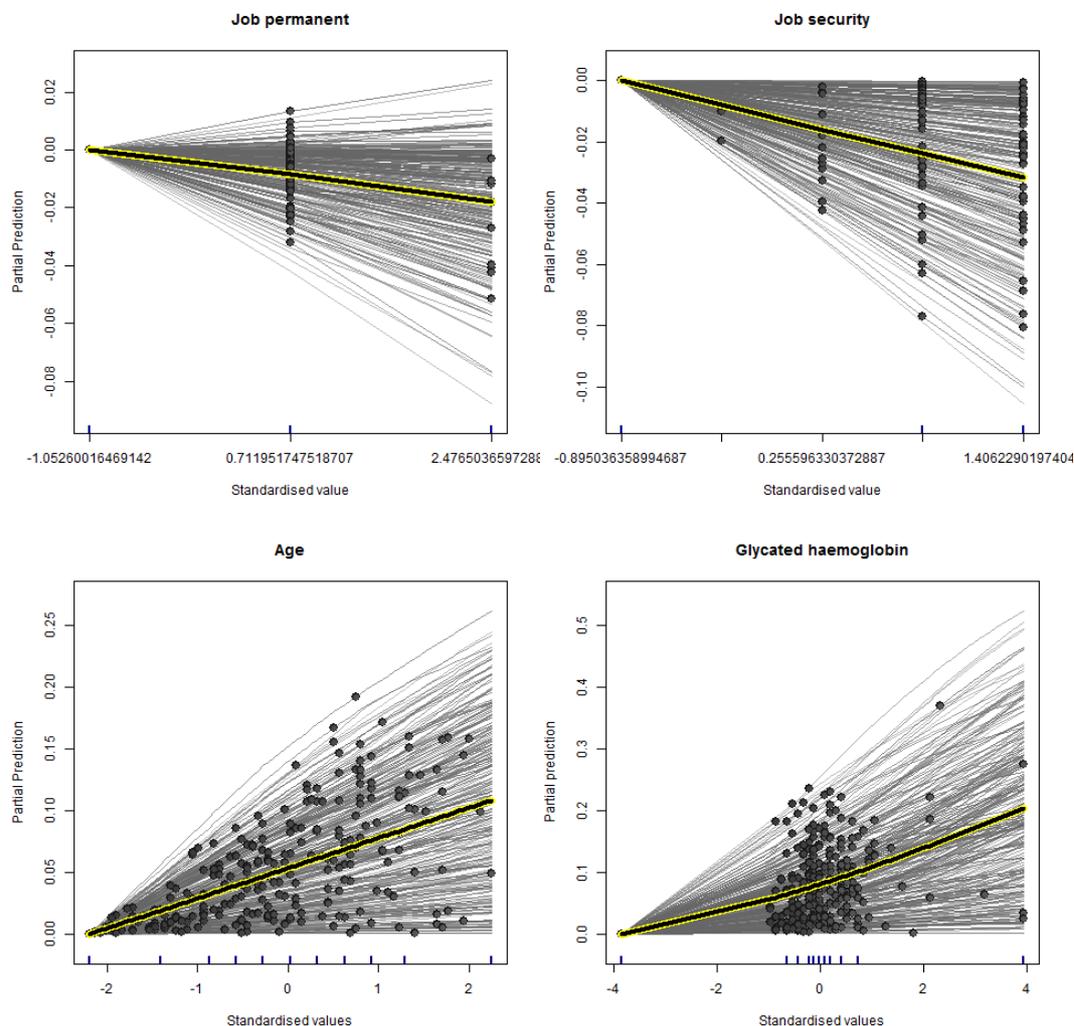

**Figure C1: ICE plots for selected variables from the neural network predicting limiting long-term illness one year from baseline.**

Figure C1 shows the results for model 1. Whether an individual's job is permanent (1 yes) or not (2 no) shows a negative relationship whereby individuals not employed in permanent jobs were given a lower probability of ill health. While this may be contrary to hypothesised directions, it likely reflects the concentration of younger (i.e. healthier) adults in insecure employment. A similar negative association was also found for job security (1 very likely, 2 likely, 3 unlikely, 4 very unlikely). The two features displaying positive associations with fairly large partial prediction values suggesting they were important (compare sizes to top two variables). Age has a positive association whereby older individuals had a higher probability of receiving a larger prediction. Higher values of glycated haemoglobin (hba1c) were also associated with poorer health which is expected.

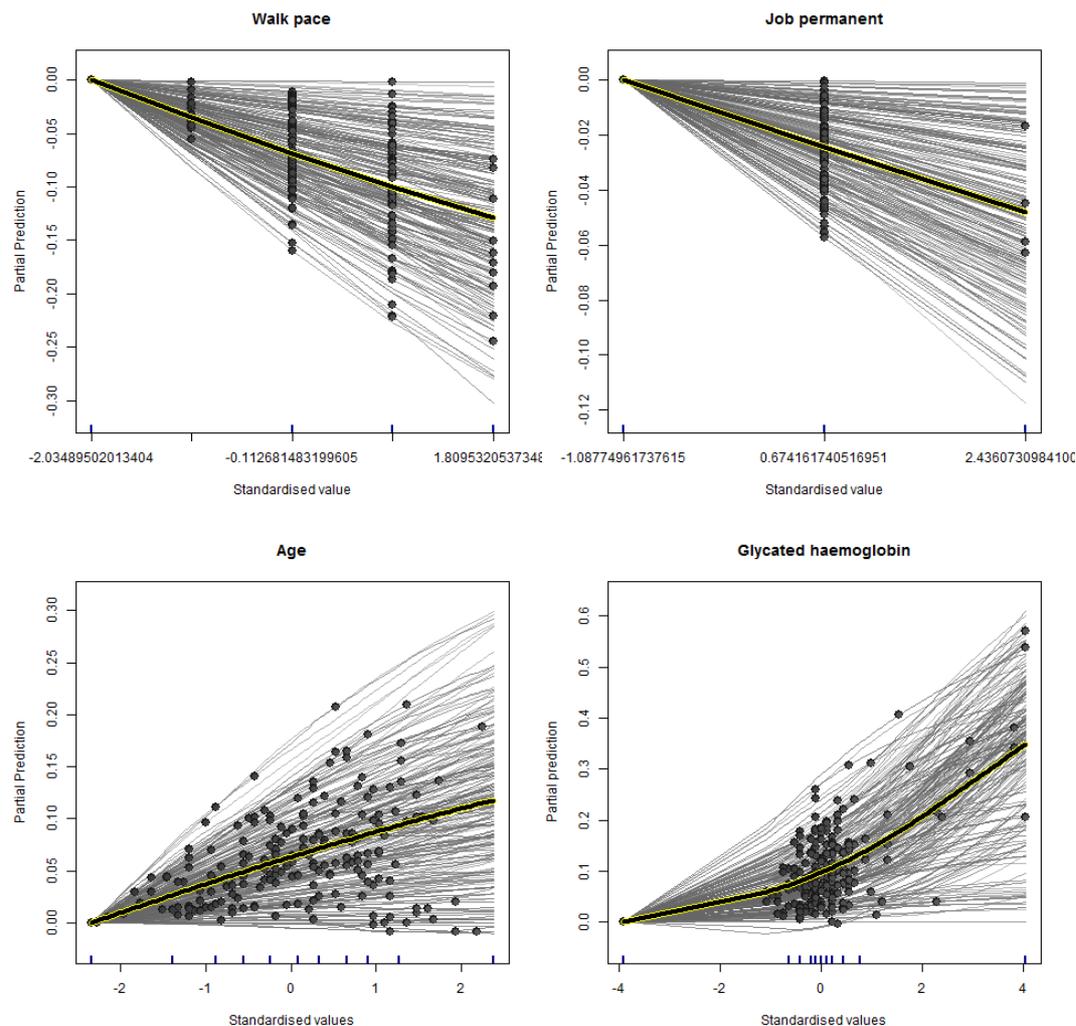

**Figure C2: ICE plots for selected variables from the neural network predicting limiting long-term illness five years from baseline.**

Figure C2 presents the analyses from model 2. Three of the same variables from the previous plot (job permanent, glycated haemoglobin and age) were present again and displayed similar associations to those depicted in Figure C1 (although note that glycated haemoglobin has a slight exponential association here). Walking pace (1 does not walk, 2 slow pace, 3 average pace, 4 brisk pace, 5 fast pace) displayed a negative association whereby individuals who walked faster were predicted lower likelihood of ill health. The partial prediction values for walking pace are also fairly large as well suggesting it was a useful predictor.

Figure C3 presents the results from model 3. Whether an individual receives state pension (1 yes, 2 no) was slightly negatively associated to predictions. It may be expected that older adults receiving pensions would have poorer health, however the results suggests the diversity of health experiences in younger than retirement adults. Physicality of work (1 very, 2 fairly, 3 not very, 4 not at all) was also negatively associated and suggested that individuals employed in occupations with less physical activities received lower predicted values. Many lower paying manual occupations have physical elements to employment, compared to higher paying managerial positions. The effect for age is now only slightly positive and there is clear diversity in predicted curves (i.e. some are positive as expected, but others negative). It suggests it was not a useful predictor of short term changes in ill health. Finally, marital status (1 single, 2 married, 3 separated, 4 widowed) has a small positive association and suggests individuals who experienced partnership dissolution were associated with poorer health.

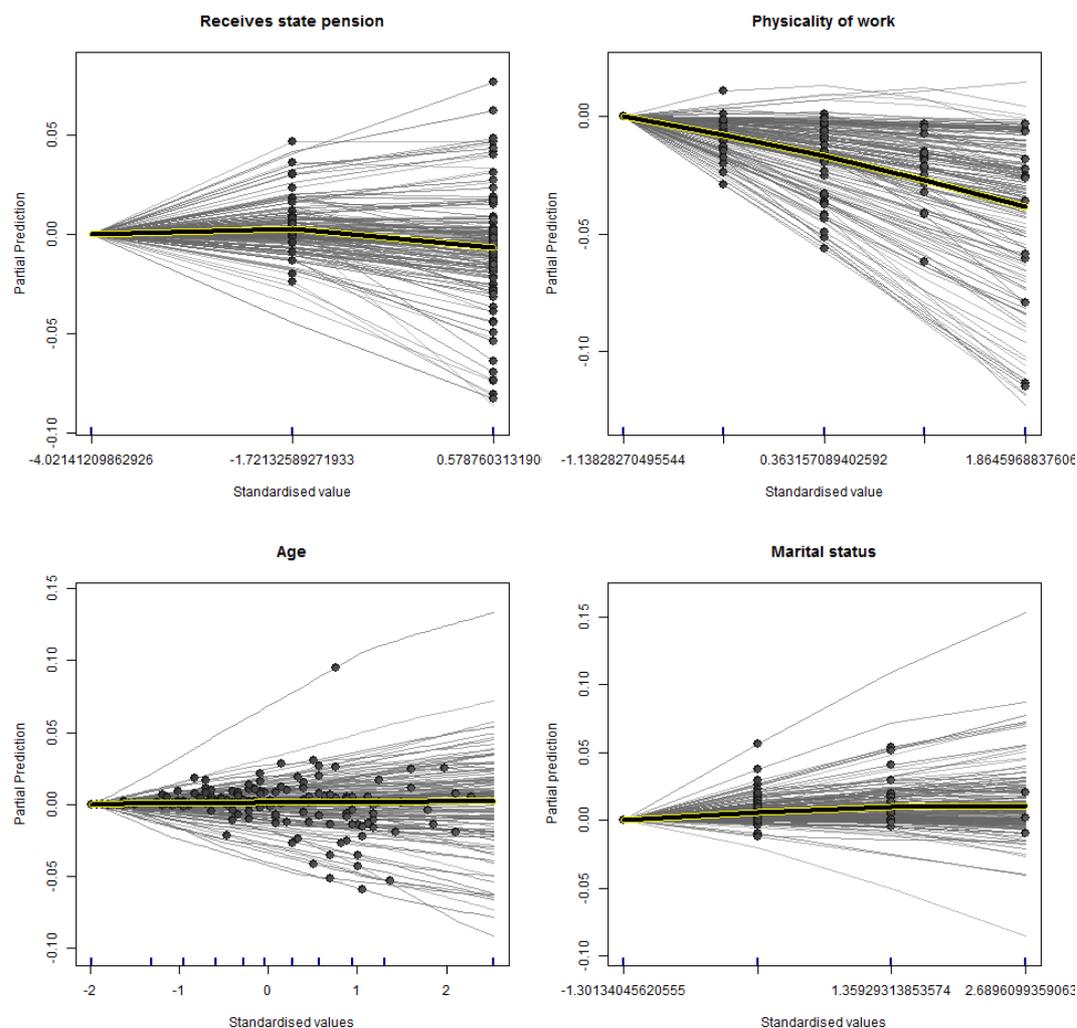

Figure C3: ICE plots for selected variables from the neural network predicting limiting long-term illness one year from baseline (excluding individuals with pre-existing limiting long-term illness at baseline).

Finally, Figure C4 presents the results for model 4. The two selected variables displaying negative relationships were for biomarkers. Insulin-like-growth factor 1 is a hormone associated with growth and anabolic processes. Dihydroepiandrosterone sulfate represents steroid hormones in the body that help control body function and repair following illness. Lower values of both biomarkers would be hypothesised to be associated with poorer health. The ICE plots demonstrate that larger values of both biomarkers are associated with lower predictions (i.e. lower risk of poor health). Whether an individual received child benefits (1 yes, 2 no) was positively associated to poor health and likely reflects individuals who are eligible being younger than those who had aged (e.g. grandparents). The association for age follows a similar pattern reported above.

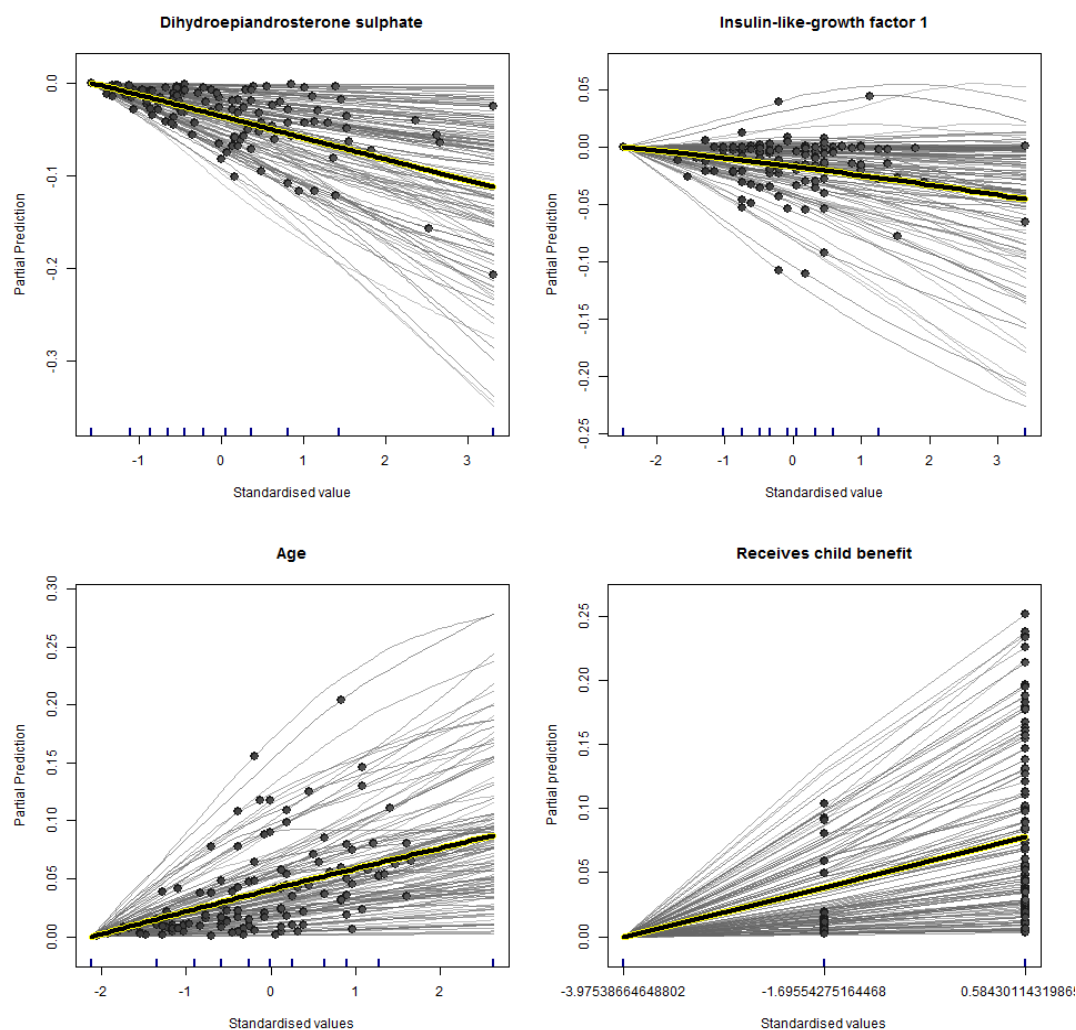

**Figure C4: ICE plots for selected variables from the neural network predicting limiting long-term illness five years from baseline (excluding individuals with pre-existing limiting long-term illness at baseline).**

# Appendix D – XGBoost model interpretation

Partial dependency plots (PDPs) were calculated to examine and visualise the nature of relationships between the predictors and outcome. They calculate the marginal effect of a predictor whilst holding the effects of all other predictors constant. PDPs were calculated for the top four most important variables within each XGBoost model (see Figure 2).

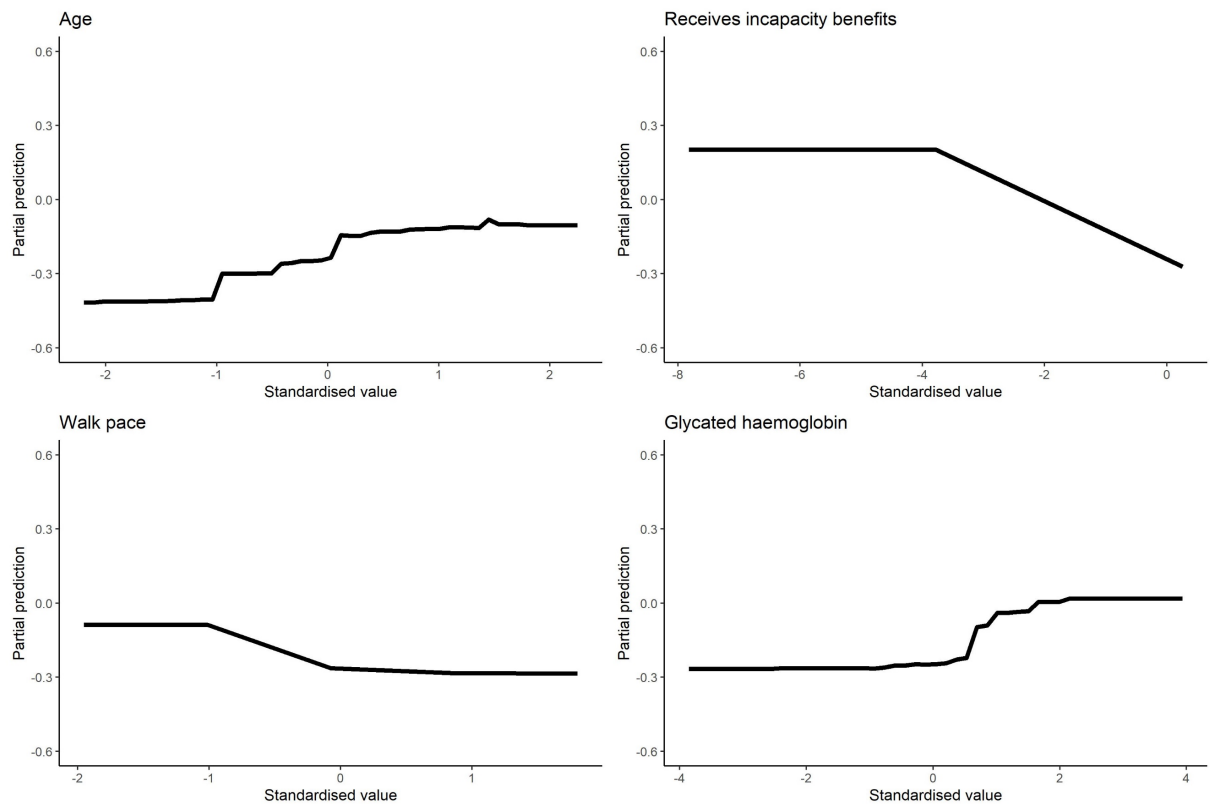

**Figure D1: Partial dependency plots for the four most important predictors from the XGBoost model predicting limiting long-term illness one year from baseline.**

Figure D1 presents the results from model 1, with results falling in expected directions. Age demonstrated a positive association with older adults receiving higher predictions of ill health. Individuals who were in receipt of incapacity benefits (1 yes, 2 no) had higher probability of ill health than compared to individuals without. Individuals with faster walking paces (1 does not walk, 2 slow pace, 3 average pace, 4 brisk pace, 5 fast pace) were predicted lower probability of ill health. Higher values of glycated haemoglobin (hba1c) were also associated with poorer health which is expected, albeit the association demonstrates a clear step change rather than a linear or general association.

Figure D2 presents the results from model 2, which contained two of the same predictors (age and walk pace) displaying similar associations to those described above. Arthritis was also included here, with individuals who had arthritis (1 yes, 2 no) having poorer future health. Individuals who felt more commonly depressed about their job (1 never, 2 occasionally, 3 some of the time, 4 most of the time, 5 all of the time) were associated with poorer future health (albeit it appears for 5)

Figure D3 presents the results for model 3. Associations for age and glycated haemoglobin were similar as those reported previously. A negative association was found for Dihydroepiandrosterone sulphate which follows hypothesised expectations (larger values of the biomarker are associated with better health). A positive association was detected for triglycerides as expected also (higher values represent greater fats in the blood and are associated with poorer health).

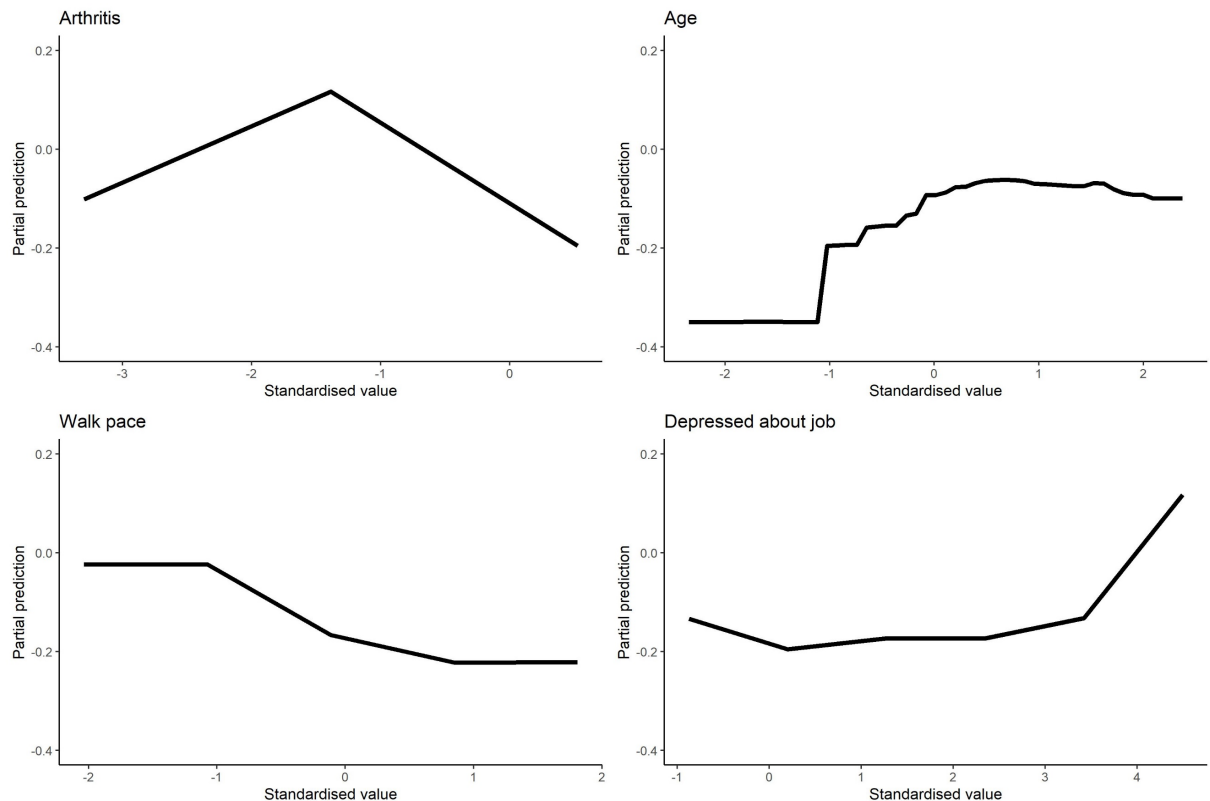

**Figure D2: Partial dependency plots for the four most important predictors from the XGBoost model predicting limiting long-term illness five years from baseline.**

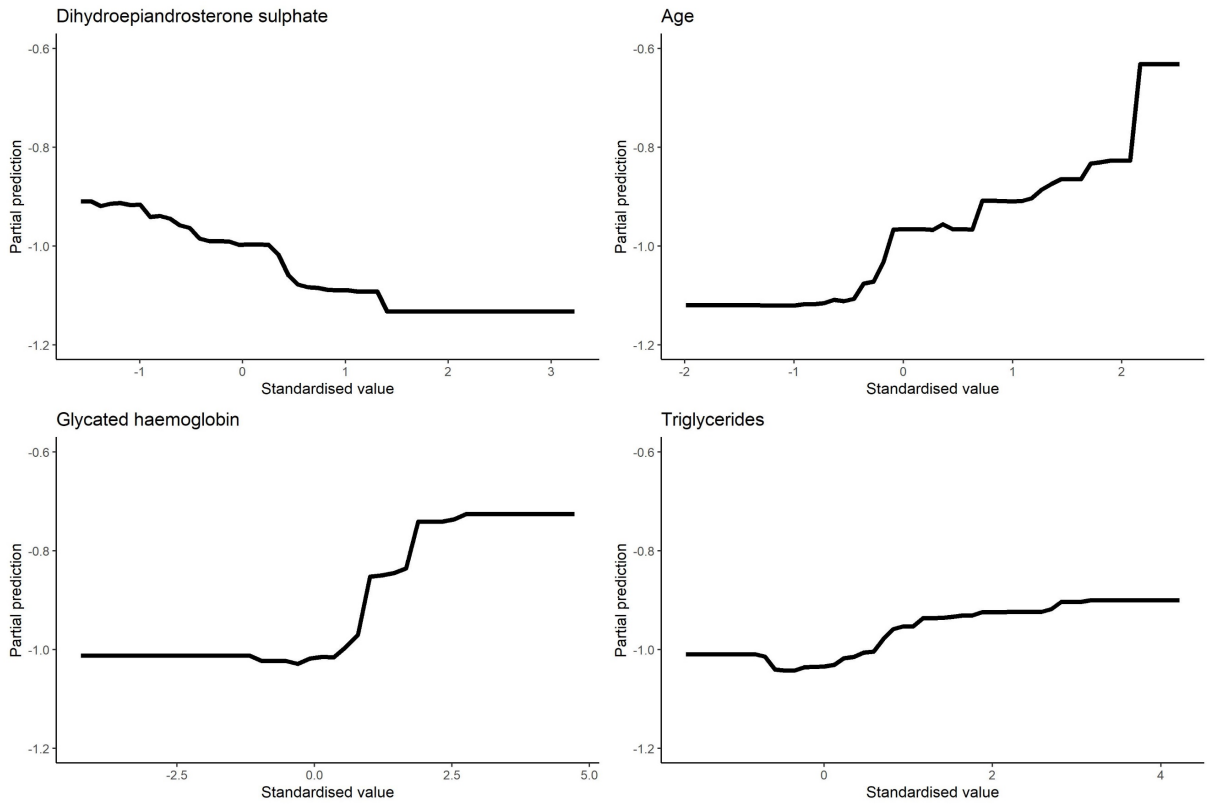

**Figure D3: Partial dependency plots for the four most important predictors from the XGBoost model predicting limiting long-term illness one year from baseline (excluding individuals with pre-existing limiting long-term illness at baseline).**

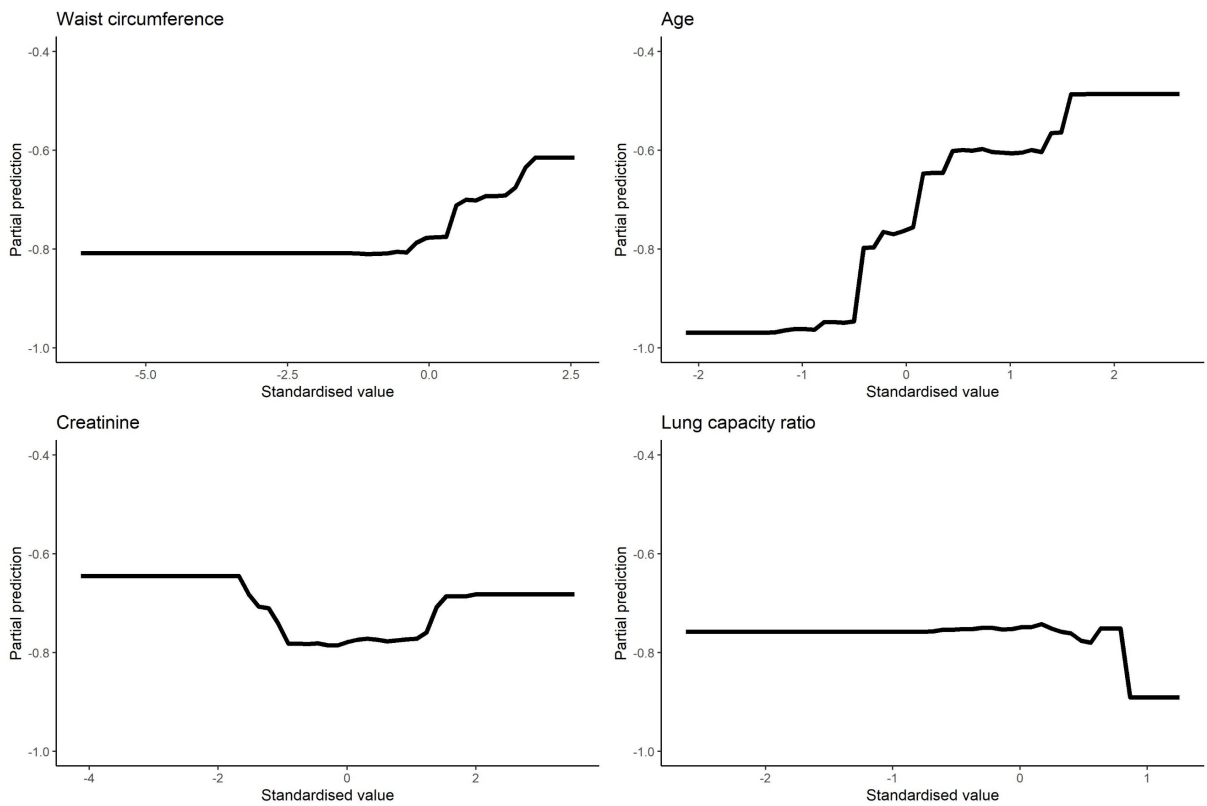

**Figure D4: Partial dependency plots for the four most important predictors from the XGBoost model predicting limiting long-term illness five years from baseline (excluding individuals with pre-existing limiting long-term illness at baseline).**

Finally, Figure D4 presents the results for model 4. The associations for age are similar to as before (in the expected direction). Lung capacity ratio (ration of forced expiratory volume in one second to forced vital capacity) displays a stepped-changed relationship with higher values associated with better health. This follows expected relationships, as a higher ratio represents better lung functioning. Waist circumference displays a positive association which follows the expected direction (i.e. individuals with larger waists are associated with poorer health). Creatinine, a measure of kidney functioning, displays an 'u-shaped relationship' that cannot be explained. Lower values would be expected to be associated with poorer kidney functioning and health.